# DIMM-SC: A Dirichlet mixture model for clustering droplet-based single cell transcriptomic data

Zhe Sun[1], Ting Wang[2], Ke Deng[3], Xiao-Feng Wang[4], Robert Lafyatis[5], Ying Ding[1], Ming Hu[4,*], Wei Chen[1,2,*]

[1]Department of Biostatistics, University of Pittsburgh Graduate School of Public Health, Pittsburgh, Pennsylvania, USA, [2]Division of Pulmonary Medicine, Allergy and Immunology; Department of Pediatrics, Children's Hospital of Pittsburgh of UPMC, University of Pittsburgh, Pittsburgh, Pennsylvania, USA, [3]Center for Statistical Science, Tsinghua University, Beijing, China, [4]Department of Quantitative Health Sciences, Lerner Research Institute, Cleveland Clinic Foundation, Cleveland, Ohio, USA, [5]Department of Medicine, Division of Rheumatology and Clinical Immunology, University of Pittsburgh School of Medicine, Pittsburgh, Pennsylvania, USA

## Abstract

**Motivation:** Single cell transcriptome sequencing (scRNA-Seq) has become a revolutionary tool to study cellular and molecular processes at single cell resolution. Among existing technologies, the recently developed droplet-based platform enables efficient parallel processing of thousands of single cells with direct counting of transcript copies using Unique Molecular Identifier (UMI). Despite the technology advances, statistical methods and computational tools are still lacking for analyzing droplet-based scRNA-Seq data. Particularly, model-based approaches for clustering large-scale single cell transcriptomic data are still under-explored.

**Methods:** We developed DIMM-SC, a **Di**richlet **M**ixture **M**odel for clustering droplet-based **S**ingle **C**ell transcriptomic data. This approach explicitly models UMI count data from scRNA-Seq experiments and characterizes variations across different cell clusters via a Dirichlet mixture prior. An expectation-maximization algorithm is used for parameter inference.

**Results:** We performed comprehensive simulations to evaluate DIMM-SC and compared it with existing clustering methods such as K-means, CellTree and Seurat. In addition, we analyzed public scRNA-Seq datasets with known cluster labels and in-house scRNA-Seq datasets from a study of systemic sclerosis with prior biological knowledge to benchmark and validate DIMM-SC. Both simulation studies and real data applications demonstrated that overall, DIMM-SC achieves substantially improved clustering accuracy and much lower clustering variability compared to other existing clustering methods. More importantly, as a model-based approach, DIMM-SC is able to quantify the clustering uncertainty for each single cell, facilitating rigorous statistical inference and biological interpretations, which are typically unavailable from existing clustering methods.

**Availability:** DIMM-SC has been implemented in a user-friendly R package with a detailed tutorial available on www.pitt.edu/~wec47/singlecell.html.
**Contact:** wei.chen@chp.edu or hum@ccf.org.

## 1 Introduction

Single cell RNA sequencing (scRNA-Seq) technologies have advanced rapidly in recent years (Gawad, et al., 2016). Among them, the newly developed droplet-based technologies have generated great interests (Macosko, et al., 2015; Zheng, et al., 2017). They are able to measure the transcriptome of thousands of single cells simultaneously in a short time period and at a relatively low cost (Macosko, et al., 2015; Zheng, et al., 2016). More attractively, droplet-based technologies utilize Unique Molecular Identifier (UMI) to annotate the 3' end of each transcript in order to reduced PCR amplification bias, increase transcript capture efficiency, and substantially minimize batch effect (Islam, et al., 2014; Kivioja, et al., 2012). More recently, 10X Genomics has released a commercialized droplet-based Chromium system, which is efficient and cost-effective in isolating thousands of single cells with average running time of ten minutes based on the Gel bead in Emulsion (GEM) technology. They used this platform to comprehensively characterize and profile peripheral blood mononuclear cells (PBMC) (Zheng, et al., 2017). Harnessing the power of these exciting new technologies, droplet-based scRNA-Seq has brought revolutionary insights to understand cellular and molecular processes at single cell resolution.

One important question in the analysis of scRNA-Seq data is to identify and characterize cell subtypes from heterogeneous tissues, which is essential to fully understand cell identity and cell function. Clustering methods have been extensively studied for many areas in the past decades. For example, unsupervised clustering methods such as K-means clustering, hierarchical clustering, and Adaptive Density Peak (ADP) clustering (Rodriguez and Laio, 2014; Wang and Xu, 2015), can be applied to droplet-based scRNA-Seq data after certain data transformation. In addition, tailored methods such CellTree and Seurat have been proposed to analyze scRNA-Seq data with the motivation from early generation platforms (duVerle, et al., 2016; Jaitin, et al., 2014). However, clustering methods tailored to droplet-based scRNA-Seq data are largely lagging behind. Although existing clustering methods can be adapted, there are at least three key limitations of using those methods to cluster droplet-based scRNA-Seq data. First of all, most existing methods are developed for continuous data (e.g. Fragments Per Kilobase of transcript per Million (FPKM) or log-transformed count data) while droplet-based scRNA-Seq data consist of the discrete count of the unique UMIs, which are direct measurements of transcript copies from each gene. Converting UMI counts into continuous measure will alter the straightforward interpretation of UMI, thus it is more appealing and reasonable to directly model the count data. Second, most existing methods are designed for the early generation of scRNA-Seq technologies that measure transcriptome across a relatively small number of single cells. It is unclear how these methods can be scaled up to cluster droplet-based scRNA-Seq data, which usually contain thousands of single cells. Last but not the least, most existing methods only provide a "hard" cluster membership for each cell without statistical uncertainty quantification. In order to conduct rigorous statistical inference and achieve reliable data interpretation, different sources of uncertainties in droplet-based scRNA-Seq data need to be explicitly taken into consideration in the clustering analysis.

To fill in these gaps, we proposed DIMM-SC, a **Di**richlet **m**ixture **m**odel for clustering droplet-based **sc**RNA-Seq data. DIMM-SC explicitly models both the within-cluster and between-cluster variability of the UMI count data, leading to rigorous quantification of clustering uncertainty for each single cell. We also implemented an efficient expectation-maximization (E-M) algorithm (Dempster, et al., 1977) for fast convergence. Furthermore, we proposed different strategies for initial value selection to ensure algorithm robustness. In the following sections, we first introduce the unique features of droplet-based scRNA-Seq data, as well as the details of the DIMM-SC method. Next, we compare the performance of DIMM-SC with three popular clustering methods, including K-means clustering, CellTree and Seurat, in both simulation studies and real data applications. K-means is one of the most popular clustering methods and has been used in the first 10X genomics publication (Zheng, et al., 2017). CellTree has been recently developed to cluster scRNA-Seq data based on Latent Dirichlet Allocation (LDA) (duVerle, et al., 2016). Seurat is a deterministic approach which relies on a graph-based clustering approach (Satija, et al., 2015).

## 2 Methods

### 2.1 Data description

The droplet-based scRNA-Seq data can be summarized into a UMI count matrix (**Table 1**), in which each row represents one gene and each column represents one single cell. Each entry in the UMI count matrix is the number of transcripts (unique UMIs) for one gene in one single cell. Compared to the data generated from early generation of scRNA-Seq technologies, droplet-based scRNA-Seq data have three important features (Gawad, et al., 2016; Stegle, et al., 2015; Zheng, et al., 2017). First, each experiment can generate thousands of cells, which dramatically increase the data dimension and computational burden. Second, the use of UMI can reduce PCR amplification bias and quantify the copies of captured molecules. Droplet-based sequencing protocol amplifies the 3' end of the transcript, so the number of UMI is independent of the total transcript length. The normalization method used in RPKM and FPKM, which adjusts for total transcript length, is invalid for analyzing droplet-based scRNA-Seq data. Therefore, the raw count data should be directly modeled to retain their biological interpretations. Third, the UMI count matrix is extremely sparse, and thus violates the statistical assumption of many existing clustering methods. Pre-selection of informative single cells and informative genes are necessary before the downstream clustering analysis. After clustering analysis, the results are usually visualized by a t-distributed stochastic neighbor embedding (t-SNE) approach (van der Maaten and Hinton, 2008), which embeds high-dimensional transcriptome data into a two-dimensional scatter plot. Note that t-SNE is a visualization tool, and it is not intended to be used for clustering scRNA-Seq data.

**Table 1.** An example of the raw UMI count table from droplet-based scRNA-Seq data

| | **Cell 1** | **Cell 2** | **Cell 3** | **…** | **Cell 2,000** |
|---|---|---|---|---|---|
| **Gene1** | 0 | 0 | 0 | … | 0 |
| **Gene2** | 1 | 0 | 1 | … | 0 |
| **Gene3** | 23 | 12 | 9 | … | 3 |
| **…** | … | … | … | … | … |
| **Gene 10,000** | 22 | 6 | 7 | 9 | 3 |

### 2.2 Statistical model

We start with a matrix $\boldsymbol{X}$, of which the element $X_{ij}$ represents the number of unique UMIs for gene $i$ in cell $j$ where $i$ runs from 1 to the total number of genes $G$, and $j$ runs from 1 to the total number of cells $C$ (as showed in **Table 1**). $X_{ij}$ is the count for the absolute number of transcripts. We denote the $j$ th column of this matrix, which gives the number of unique UMIs in the $j$ th single cell, by a vector $\boldsymbol{x_j} = (x_{1j}, x_{2j}, \dots, x_{Gj})$, where $j = 1, \dots, C$. We assume that $\boldsymbol{x_j}$ is generated from a multinomial distribution with parameter vector $\boldsymbol{p_j} = (p_{1j}, p_{2j}, \dots, p_{Gj})$. The element of $\boldsymbol{p_j}$, $p_{ij}$, is the probability that a unique UMI count taken from cell $j$ belongs to gene $i$. This gives a likelihood for each cell:

$$P(\boldsymbol{x_j}|\boldsymbol{p_j}) = \frac{T_j!}{\prod_{i=1}^{G} x_{ij}!} p_{1j}^{x_{1j}} p_{2j}^{x_{2j}} \dots p_{Gj}^{x_{Gj}},$$

where $T_j = \sum_i x_{ij}$ is the total number of unique UMIs for the $j$ th cell. The joint likelihood of all $C$ cells is the product of the likelihood for each cell: $\prod_{j=1}^{C} P(\boldsymbol{x_j}|\boldsymbol{p_j})$.

#### 2.2.1 Dirichlet mixture priors

In a Bayesian framework, we need to define a prior distribution for the multinomial parameter probability vector $\boldsymbol{p_j}$. For multinomial distribution, a commonly-used conjugate prior is the Dirichlet distribution. Specifically, we assume that the proportion $\boldsymbol{p_j} = (p_{1j}, p_{2j}, \dots, p_{Gj})$ follows a Dirichlet prior distribution $Dir(\boldsymbol{\alpha}) = Dir(\alpha_1, \alpha_2, \dots, \alpha_G)$:

$$P(\boldsymbol{p_j}|\boldsymbol{\alpha}) = \frac{1}{B(\boldsymbol{\alpha})} p_{1j}^{\alpha_1 - 1} p_{2j}^{\alpha_2 - 1} \dots p_{Gj}^{\alpha_G - 1},$$

where $B(\boldsymbol{\alpha})$ is Beta function with parameter $\boldsymbol{\alpha} = (\alpha_1, \alpha_2, \dots, \alpha_G)$. All the elements in $\boldsymbol{\alpha}$ are strictly positive ($\alpha_i > 0$). The mean and variance of $p_{ij}$ are $\alpha_i/|\boldsymbol{\alpha}|$ and $\alpha_i(|\boldsymbol{\alpha}| - \alpha_i)/(|\boldsymbol{\alpha}|^2(|\boldsymbol{\alpha}| + 1))$, respectively, where $|\boldsymbol{\alpha}| = \alpha_1 + \alpha_2 + \dots + \alpha_G$. A large $|\boldsymbol{\alpha}|$ gives small variance about the proportions $\boldsymbol{p_j}$, while a small $|\boldsymbol{\alpha}|$ leads to widely spread $\boldsymbol{p_j}$'s. When the cell population is homogeneous, we assume that $\boldsymbol{p_j}$'s all follow the same prior distribution $Dir(\boldsymbol{\alpha})$, and the full likelihood function is as follows:

$$P(\boldsymbol{x_j}|\boldsymbol{\alpha}) = \int P(\boldsymbol{x_j}|\boldsymbol{p_j})P(\boldsymbol{p_j}|\boldsymbol{\alpha})\, d\boldsymbol{p_j} = \frac{T_j!}{\prod_{i=1}^{G} x_{ij}!}\left(\prod_{i=1}^{G}\frac{\Gamma(x_{ij}+\alpha_i)}{\Gamma(\alpha_i)}\right)\frac{\Gamma(|\boldsymbol{\alpha}|)}{\Gamma(T_j+|\boldsymbol{\alpha}|)}.$$

We then assume that the cell population consists of $K$ distinct cell types, where $K$ can be pre-defined according to prior biological knowledge or can be estimated through model fitting. To provide a more flexible modeling framework and allow for unsupervised clustering, we extend the aforementioned single Dirichlet prior to a mixture of $K$ Dirichlet distributions, indexed with $k = 1, \dots, K$, each with parameter $\boldsymbol{\alpha}_{(k)}$. If cell $j$ belongs to the $k$ th cell type, its gene expression profile $\boldsymbol{p_j}$ follows a cell-type-specific prior distribution $Dir(\boldsymbol{\alpha}_{(k)})$. The full likelihood function is then obtained by multiplying the Dirichlet mixture prior by the multinomial likelihood.

### 2.2.2 E-M algorithm for fitting the mixture of Dirichlet prior

We now use a latent variable vector $\boldsymbol{Z}$ with elements $z_j$ to represent the cell type label for the cell $j$. This allows us to maximize the log posterior distribution using the E-M algorithm (Dempster, 1977). We have

$$P(\boldsymbol{x_j}|z_j = k, \boldsymbol{\alpha}_{(k)}) \propto \left(\prod_{i=1}^{G}\frac{\Gamma(x_{ij}+\alpha_{ik})}{\Gamma(\alpha_{ik})}\right)\frac{\Gamma(|\boldsymbol{\alpha}_{(k)}|)}{\Gamma(T_j+|\boldsymbol{\alpha}_{(k)}|)},$$

and $P(z_j = k) = \pi_k$, where $\pi_k$ is the proportion of the $k$ th cell type among all cells. We can treat $z_j$ as missing data, and use the E-M algorithm to estimate $\alpha_{1k}, \alpha_{2k}, \dots, \alpha_{Gk}$ and $\pi_k$. The complete log likelihood is

$$\log \prod_{j=1}^{C} P(\boldsymbol{x_j}, z_j = k) = \sum_{j=1}^{C} I(z_j = k) \log\left\{\left(\prod_{i=1}^{G}\frac{\Gamma(x_{ij}+\alpha_{ik})}{\Gamma(\alpha_{ik})}\right)\frac{\Gamma(|\boldsymbol{\alpha}_{(k)}|)}{\Gamma(T_j+|\boldsymbol{\alpha}_{(k)}|)}\right\}.$$

The formula for updating $\alpha_{1k}, \alpha_{2k}, \dots, \alpha_{Gk}$ is derived from the Minka's fixed-point iteration for the leaving-one-out likelihood (Minka, 2000):

$$\hat{\alpha}_{ik}^{(t+1)} = \alpha_{ik}^{(t)}\frac{\sum_{j=1}^{C}\delta_{jk}\{x_{ij}/(x_{ij}-1+\alpha_{ik}^{(t)})\}}{\sum_{j=1}^{C}\delta_{jk}\{T_j/(T_j-1+|\boldsymbol{\alpha}_{(k)}^{(t)}|)\}}.$$

We repeat the above steps until the convergence of log likelihood or a maximum number of iterations is reached (see detailed algorithm in supplemental materials).

### 2.2.3 Selection of the number of clusters and initial values

To implement DIMM-SC, it is critical to select the total number of clusters and the initial values for the E-M algorithm. Specially, the number of clusters $K$ can be defined with prior knowledge or can be selected from model selection criteria such as AIC or BIC (Akaike, 1974; Schwarz, 1978). Meanwhile, there are many methods to determine the initial values of $\alpha_1, \alpha_2, \dots, \alpha_G$ in the E-M algorithm for fitting the Dirichlet mixture model. For example, Ronning (1989) suggests to estimate $\sum_{i=1}^{G}\alpha_i$ by

$$\log \sum_{i=1}^{G}\alpha_i = \frac{1}{G-1}\sum_{i=1}^{G-1} log\left(\frac{E(p_i)(1-E(p_i))}{var\,(p_i)} - 1\right),$$

where $E(p_i)$ can be approximated by $(\sum_{j=1}^{C} x_{ij}/\,T_j)/C)$ (Ronning, 1989). An alternative approach is to estimate the initial values using a method of moment estimates proposed by Weir and Hill (Weir and Hill, 2002). In this work, we applied K-means or ADPclust to get a preliminary clustering result, and then used either the Ronning's method or the Weir and Hill's method to estimate initial values of $\boldsymbol{\alpha}$.

## 2.3 Simulation studies

We performed comprehensive simulation studies to compare DIMM-SC with three existing clustering methods, including K-means clustering, Seurat and CellTree. The first two are deterministic approaches and the third one is a probabilistic approach.

In the simulation set-up, the UMI count matrix was sampled from the proposed Dirichlet mixture model. Specially, for a fixed total number of cell clusters $K$, we first pre-defined the values of $\boldsymbol{\alpha}_{(k)} = (\alpha_{1(k)}, \alpha_{2(k)}, \dots, \alpha_{G(k)})$ for the $k$ th cell cluster, and then sampled the proportion $\boldsymbol{p_j} = (p_{1j}, p_{2j}, \dots, p_{Gj})$ from a Dirichlet distribution $Dir(\boldsymbol{\alpha}_{(k)})$. Next, we sampled the UMI count vector $\boldsymbol{x_j}$ for the $j$ th cell from the multinomial distribution $Multinomial(T_j, \boldsymbol{p_j})$. We fixed $T_j$ as a constant across all cells.

In the simulation studies, we considered the following seven clustering methods. (1) DIMM-SC + K-means + Ronning (hereafter referred as DIMM-SC-KR), in which we used the K-means clustering to obtain the initial values of clustering labels and then used the Ronning's method to estimate initial values of $\boldsymbol{\alpha}$; (2) DIMM-SC + K-means + Weir (hereafter referred as DIMM-SC-KW), in which we used the K-means clustering to obtain the initial values of clustering labels and used the Weir and Hill's method to estimate initial values of $\boldsymbol{\alpha}$; (3) DIMM-SC + random + Ronning (hereafter referred as DIMM-SC-RR), in which we randomly selected the initial values of clustering labels and used the Ronning's method to estimate initial values of $\boldsymbol{\alpha}$; (4) DIMM-SC + random + Weir (hereafter referred as DIMM-SC-RW), in which we randomly selected the initial values of clustering labels and used the Weir and Hill's method to estimate initial values of $\boldsymbol{\alpha}$; (5) K-meaning clustering; (6) CellTree, a LDA-based approach to cluster scRNA-Seq data; and (7) Seurat. To perform the simulation analysis using Seurat, we followed the tutorial instructions from the Seurat website and used all genes as input to perform Principal Component Analysis (PCA). After that, we followed the "jackstraw" procedure implemented in Seurat, and identified first ten PCs for their downstream algorithm. We fixed the number of PCs in all the simulation runs under each scenario. Since Seurat requires users to self-specify a resolution parameter with increased values leading to a greater number of clusters, the clustering results are very sensitive to this resolution parameter.

Seurat suggests that setting this resolution parameter between 0.6-1.2 typically returns good results for datasets of around 3,000 cells, so we ran Seurat using resolution parameter with 0.6, 0.8, 1.0 and 1.2, and chose the one with the highest ARI value in each simulation setting.

We used the signal-to-noise ratio (SNR) to measure the magnitude of difference among different cell clusters. When $K = 2$, SNR is defined as:

$$SNR = \frac{|\boldsymbol{\alpha}_{(1)} - \boldsymbol{\alpha}_{(2)}|_1}{G\sqrt{Var(\boldsymbol{\alpha}_{(1)}) + var(\boldsymbol{\alpha}_{(2)})}},$$

where $|.|_1$ is the $L_1$ norm of a vector. We performed comprehensive simulations to investigate how different SNRs, different sequencing depths, different total numbers of cells/genes/clusters, and different proportions of noisy genes affect the clustering results. To evaluate the performance of DIMM-SC and other competing clustering methods, we used the following two metrics: (1) clustering accuracy measured by the adjusted rand index (ARI) (Rand, 1971), which is a metric of the similarity between the estimated clustering labels and the true clustering labels and (2) Stability (the standard deviation of ARI). We expect a good clustering method should achieve both high accuracy and high stability.

## 3 Results

### 3.1 Simulation studies

**Figure 1A** shows the boxplots of ARI for seven clustering methods across 100 simulations at different SNRs. Four DIMM-SC based methods (KR, KW, RR, RW) achieved comparable performance, which produced higher accuracy and lower variability than K-means clustering, Seurat and CellTree. When SNR is high (i.e., substantial differences among cell clusters), all seven methods performed well. However, when SNR is low (i.e., different cell clusters are similar), K-means clustering, Seurat and CellTree produced less accurate and more variable clustering results, while four DIMM-SC based methods still performed well.

**Figure 1B** shows the boxplots of ARI for seven clustering methods across 100 simulations, when the total number of clusters is 2, 3 and 4, respectively. The four DIMM-SC based methods, especially the two methods with randomly selected initial cluster labels (RR and RW), achieved better clustering accuracy (i.e., higher ARI) with more number of clusters. K-means clustering has high variability for more clusters, since it is a deterministic procedure and is more likely to end at a local optimum when the total number of clusters increases. CellTree performed worse for more clusters, partially due to the over-parameterized LDA model and lack of fit to highly heterogeneous data. Seurat was run under different default recommended parameters and the performance varies with different parameters.

**Figure 1C~F** list the boxplots of ARI for seven clustering methods across 100 simulations, for different number of genes (**Figure 1C**), different number of cells (**Figure 1D**), different sequencing depths (**Figure 1E**) and different number of informative genes (**Figure 1F**) (i.e., differentially expressed genes among clusters), respectively. Consistent across all these four scenarios, more information (i.e., more genes, more cells, higher sequencing depths and more informative genes) lead to higher clustering accuracy and lower clustering variability. Four DIMM-SC based clustering methods consistently outperformed K-means clustering, Seurat and CellTree in all these simulation settings, suggesting the advantage of DIMM-SC.

### 3.2 Real data analysis: the publicly available 10X scRNA-Seq data

#### 3.2.1 In silicon studies based on purified cell types from published scRNA-Seq data

To illustrate the application of DIMM-SC to real datasets, we first benchmarked our method against pre-defined measures in capturing true cell-to-cell similarities on published single-cell datasets. 10X Genomics has made eleven datasets from purified cell types available to public (Zheng, et al., 2017). Among which, over 10,000 cells were detected in most experiments. Here we considered two scenarios: (1) a simple case with cells from three highly distinct cell types (CD56+ NK cells, CD19+ B cells, and CD4+/CD25+ regulatory T cells); (2) a challenging case with cells from three similar cell types (CD8+/CD45RA+ naive cytotoxic T cells, CD4+/CD25+ regulatory T cells, and CD4+/CD45RA+/CD25- naive T cells) (**Table 2**). For visualization, we used the t-SNE algorithm to project the data into a two-dimensional space so that certain hidden structures in the data can be depicted intuitively (see the t-SNE visualization in **Figure S1** and **Figure S2)**.

We ran DIMM-SC, K-means clustering, CellTree and Seurat 50 times for both two scenarios. In the simple case, at each time, we randomly selected 1,000 CD56+ NK cells, 2,000 CD19+ B cells and 3,000 CD4+/CD25+ regulatory T cells from the 10X Genomics datasets, and combined them together. Thus the total number of cells for clustering is 6,000. Similarly, in the challenging case, 1,000 CD8+/CD45RA+ naive cytotoxic T cells, 2,000 CD4+/CD25+ regulatory T cells and 3,000 CD4+/CD45RA+/CD25- naive T cells were randomly selected at each time.

Cell types in each dataset were known as a priori and were further validated in the respective follow-up studies, providing a reliable gold standard to benchmark the clustering performance for each method. We compared the performance of four DIMM-SC methods with K-means clustering, CellTree and Seurat, in terms of clustering accuracy and stability.

In the simple case, we applied all these seven clustering methods on the top 100 variable genes ranked by their standard error among all cells. **Table 3** shows that all methods provided good clustering results. Two DIMM-SC methods with randomly selected initial cluster labels (RR and RW) slightly outperformed K-means clustering in terms of accuracy and variability. For the challenging case, unlike what we did in the simple case, we chose different numbers of top variable genes. **Table 3** and **Figure S3** show that the ARIs of CellTree and Seurat were lower than other methods when the total number of genes used for clustering was greater than 200. DIMM-SC outperformed K-means clustering in terms of accuracy. K-means clustering made a great leap forward when the total number of genes increased to 300. However, there is no further improvement of ARI with K-means clustering when top 500 or more variable genes were used. Since in the challenging case, CD4+/CD25+ regulatory T cells and CD4+/CD45RA+/CD25- naive T cells were similar to each other, more and more noisy genes were included in the analysis when we increased the total number of genes, which undermined the performance of K-means clustering. Note that K-means clustering and Seurat were only able to provide a deterministic clustering label, while DIMM-SC and CellTree can additionally provide the probability that each cell belongs to each cluster.

**Table 2.** Total number of cells, genes and validated populations for two scenarios, for in silicon studies based on purified cell types from 10X Genomics.

| Scenario | #Genes | #Cell | Cell type |
|---|---|---|---|
| **Simple** | 32,738 | 8,385 | CD56+ NK cells |
| | | 10,085 | CD19+ B cells |
| | | 10,283 | CD4+/CD25+ regulatory T cells |
| **Challenging** | 32,738 | 11,953 | CD8+/CD45RA+ naive cytotoxic T cells |
| | | 10,263 | CD4+/CD25+ regulatory T cells |
| | | 10,479 | CD4+/CD45RA+/CD25- naive T cells |

**Table 3.** Performance of clustering in the simple case and the challenging case

| #Genes | DIMM-SC-KR | DIMM-SC-KW | DIMM-SC-RR | DIMM-SC-RW | K-means clustering | CellTree | Seurat |
|---|---|---|---|---|---|---|---|
| | | | | The simple case | | | |
| 100 | 0.952 (0.114) | 0.951 (0.118) | 0.982 (0.052) | 0.990 (0.002) | 0.951 (0.129) | 0.983 (0.002) | 0.983 (0.003) |
| | | | | The challenging case | | | |
| 100 | 0.351 (0.140) | 0.357 (0.140) | 0.368 (0.140) | 0.408 (0.128) | 0.182 (0.012) | 0.278 (0.018) | 0.395 (0.027) |
| 200 | 0.558 (0.014) | 0.559 (0.014) | 0.558 (0.014) | 0.559 (0.013) | 0.283 (0.050) | 0.389 (0.022) | 0.410 (0.017) |
| 300 | 0.563 (0.013) | 0.564 (0.013) | 0.563 (0.013) | 0.563 (0.013) | 0.526 (0.063) | 0.419 (0.023) | 0.413 (0.022) |
| 400 | 0.571 (0.014) | 0.571 (0.014) | 0.566 (0.040) | 0.571 (0.014) | 0.554 (0.014) | 0.404 (0.050) | 0.429 (0.012) |
| 500 | 0.572 (0.015) | 0.572 (0.015) | 0.572 (0.015) | 0.572 (0.015) | 0.559 (0.014) | 0.397 (0.067) | 0.435 (0.011) |
| 800 | 0.562 (0.041) | 0.562 (0.041) | 0.557 (0.057) | 0.556 (0.056) | 0.557 (0.041) | 0.365 (0.078) | 0.445 (0.011) |

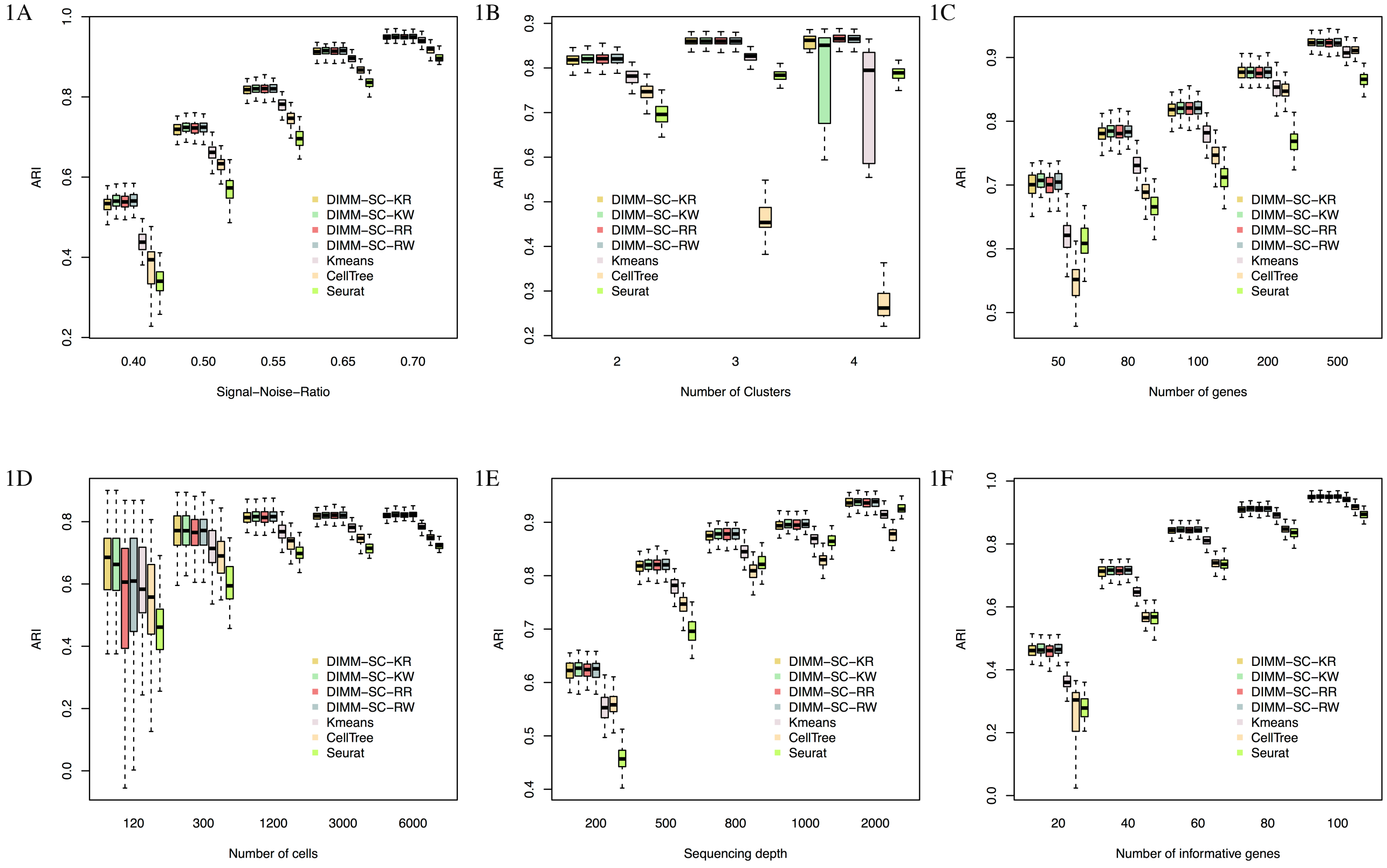

**Fig. 1.** Boxplots of ARI for seven clustering methods across 100 simulations, investigating how different SNRs (A), number of clusters (B), number of genes (C), number of cells (D), sequencing depth (E) and the number of informative genes (F) affect clustering results.

### 3.2.2 Real data analysis on the PMBC 68K dataset

To examine how DIMM-SC is applicable to large-scale dataset, we applied DIMM-SC-KR on the PBMC 68K dataset, which consists of >68,000 single cells. Among all 32,738 genes, we selected the top 1,000 genes with the highest variations. **Figure 2A** shows a clear separation of cell types as we expected. 11 purified sub-populations of PBMCs were used as the reference to identify the cell type of each single cell from the PBMC 68K dataset. We used the labels from cell classification analysis as the approximated truth. In this analysis, each cell was assigned to the purified population which has the highest correlation with its gene expression profile. We calculated ARIs between the true labels and inferred ones obtained from K-means clustering, CellTree, Seurat and DIMM-SC. The ARIs of K-means clustering, CellTree and DIMM-SC are 0.32, 0.28 and 0.41, respectively. To perform the analysis using Seurat, we used the default setting of Seurat to select the top 1,657 variable genes, and picked the first 22 PCs for the clustering analysis. The ARI of Seurat is 0.31, suggesting that DIMM-SC performed the best in the PMBC 68K dataset. Additionally, we highlighted vague cells in the t-SNE projection (**Figure 2B**), where vague cells are defined as cells with the largest posterior cluster-specific probability < 0.95. As shown in **Figure 2B**, most of vague cells are located at the boundary of different clusters, which reassuring the validity of the clustering results.

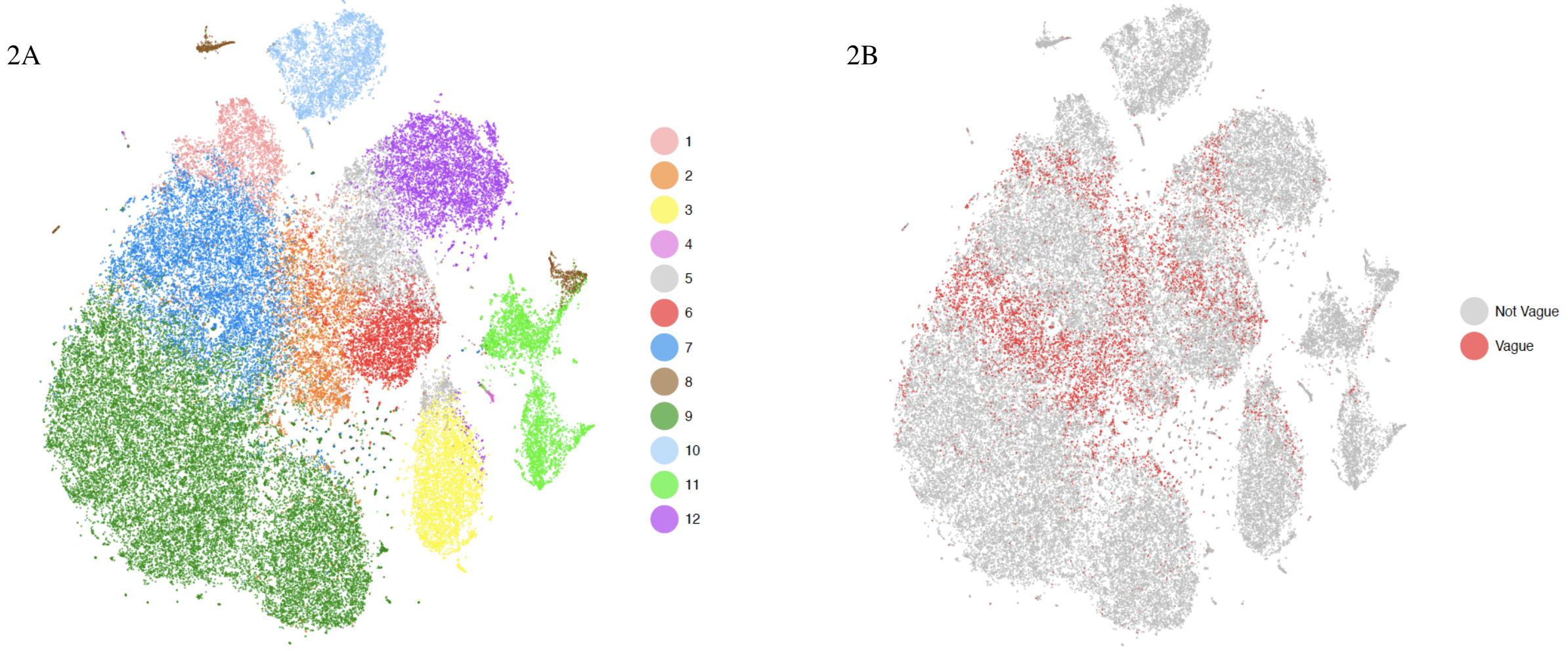

**Fig. 2.** The t-SNE projection of 68K PBMCs, colored by the DIMM-SC clustering assignment and the illustration of vague cells with the largest posterior probability < 0.95

### 3.2.3 Analysis of the in-house scRNA-Seq data from systemic sclerosis study

Collaborating with investigators at the University of Pittsburgh, we are in the first place to use the 10X Chromium system to generate scRNA-Seq data in order to study systemic sclerosis. We applied DIMM-SC to the scRNA-Seq data of skin tissue collected from a systemic sclerosis patient. Starting from a UMI count matrix for 1,180 cells generated by the 10X genomics Cellranger pipeline, we first removed cells that had less than 300 genes expressed and filtered noisy genes that were expressed in less than five cells, then we extracted the top 1,000 highly variable genes based on their standard deviations. We set the total number of clusters to be six based on our prior knowledge and utilized the KR method to generate the initial cluster labels and the initial values for the parameter $\boldsymbol{\alpha}$. The six cell clusters from DIMM-SC included 92, 89, 45, 156, 469 and 271 cells, respectively. **Figure 3** shows the t-SNE projection of the skin cells, colored by cluster labels inferred by DIMM-SC, and the dashed circles represent potential subtypes of skin cells according to the expressions of cell type specific markers. It is interesting that fibroblast cells exhibit two clusters, suggesting possible subtypes. For each cell cluster, we identified top marker genes that were differentially expressed between the specified cluster and all the other clusters. We recognized some subtypes of skin cells for the identified clusters based on the biological knowledge of cell specific markers, such as pericyte cells specifically expressed gene *RGS5*, T cells specifically expressed gene *IL32*, endothelial cells specifically expressed gene *VWF*, fibroblast cells specifically expressed gene *COL1A1*, basal keratinocyte cells specifically expressed gene *KRT14* and gene *KRT5*, and suprabasal keratinocyte cells specifically expressed gene *KRT1* and gene *KRT10*.

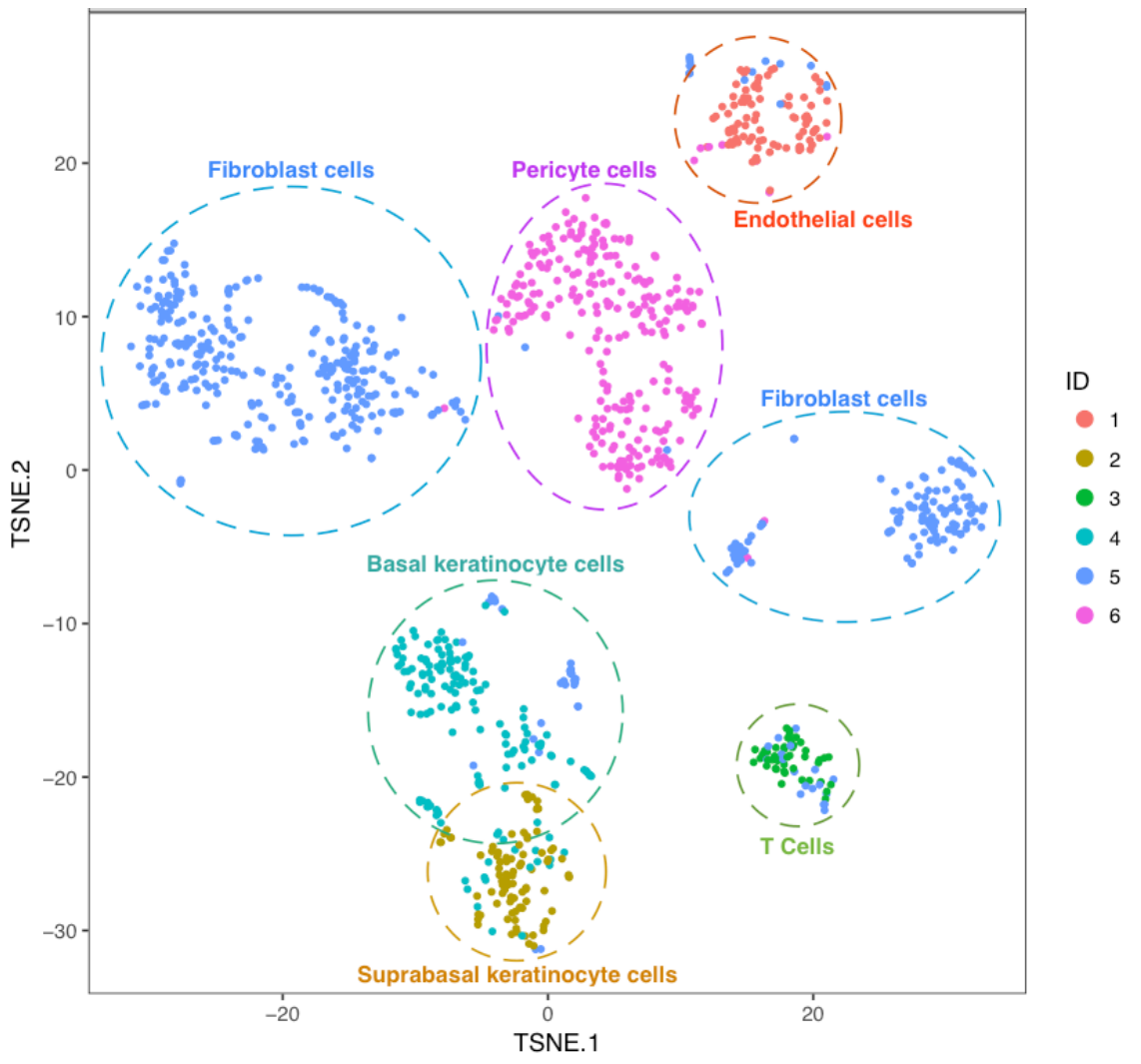

**Fig. 3.** The t-SNE projection of cells from systemic sclerosis skin tissue, colored by the DIMM-SC clustering assignment.

### 3.3 Model fitting diagnosis

An important step in applying model-based approach is to examine whether the proposed statistical model fits the real data well. In Dirichlet distribution, the marginal distribution of $\boldsymbol{p}$ is a Beta distribution. In addition, the mean of $p_i$, $\alpha_{ik}/|\boldsymbol{\alpha}_{(k)}|$, is approximately proportional to its variation $\alpha_{ik}(|\boldsymbol{\alpha}_{(k)}| - \alpha_{ik})/(|\boldsymbol{\alpha}_{(k)}|^2(|\boldsymbol{\alpha}_{(k)}| + 1))$. After applying DIMM-SC to the PBMC 68K scRNA-Seq dataset, we performed the following two analyses to evaluate the goodness of fit of the model. We first collected cells that belong to the same cell type using datasets of purified sub-populations of PBMCs from 10X Genomics, and then plotted the empirical marginal distribution of proportion $p_i$ for top variable genes. We compared such empirical distribution with the marginal distribution $Beta(\alpha_{ik}, |\boldsymbol{\alpha}_{(k)}| - \alpha_{ik})$ at $\alpha_{(k)} = \hat{\alpha}_{(k)}$, where $\hat{\alpha}_{(k)}$ was estimated from the real scRNA-Seq data. **Figure S4A** shows that the fitted distributions for top variable genes aligned very well with the empirical distributions, suggesting that DIMM-SC achieved good fit in real scRNA-Seq data.

Moreover, we explored the relationship between the mean and variance of $p_i$'s, as commonly used in count data analysis, to evaluate whether any over-dispersion pattern exists. Similar to the previous analysis, we also collected cells from the same cell type, and calculated the mean and variation of $p_i$ for each gene. The scatter plot of the log mean of $p_i$ versus the log variance of $p_i$ (**Figure S4B**) shows a clear linear relationship between mean and variance. Derived from Dirichlet distribution, the expected intercept and slope can be approximated by 1 and $\log(|\boldsymbol{\alpha}|)$, respectively, where $\log(|\hat{\boldsymbol{\alpha}}|)$ was estimated from the real scRNA-Seq data. In CD56+ Natural Killer cells and CD19+ B cells, $\log(|\hat{\boldsymbol{\alpha}}|)$ equals to 6.60 and 6.67, respectively. As shown in **Figure S4B**, the intercept and slope of the fitted line (red line) are close to the expected values, indicating a good model fitting in this real scRNA-Seq data. We noticed that, due to both technical and biological uncertainties, a few genes exhibit extra variation, which cannot be fully explained by the mean-variation relationship posited by the Dirichlet distribution. We will pursue to extend DIMM-SC to account for such additional variation in the near future.

## 4 Software availability

We have implemented DIMM-SC into a user-friendly R package, which is freely available on http://www.pitt.edu/~wec47/singlecell.html. This software program can take a full matrix file that is compiled by users or can directly take the sparse UMI count matrix file from the 10X Genomics Cellranger pipeline. The output includes clustering results, the probability matrix for all cells, the probability vector for each gene, and the t-SNE projection visualization.

## 5 Discussion

Compared with the early generation scRNA-Seq technologies, the intrinsic characteristics of droplet-based scRNA-Seq data, including a much larger number of cells and direct counting of molecule copies using UMI, pose great challenges on statistical analysis and require new methodological development. In this study, we developed a model-based clustering method DIMM-SC for analyzing droplet-based scRNA-Seq data. DIMM-SC directly models UMI counts from scRNA-Seq data using a multinomial distribution with Dirichlet mixture priors. We demonstrated that DIMM-SC has achieved substantial improvements in clustering accuracy and stability compared to existing clustering methods such as K-means clustering, Seurat and CellTree. More importantly, our probabilistic model provides clustering uncertainty for each cell (how likely each cell belongs to each cluster), thus can benefit rigorous statistical inference and straightforward biological interpretations. In addition, DIMM-SC can be used to detect differentially expressed gene markers among different cell types, which is under our further investigation.

Our probabilistic model coupled with a computationally efficient EM algorithm is able to cluster large-scale droplet-based scRNA-Seq data. For example, it takes around 3 hours to cluster 68,000 cells using top 1,000 highly variable genes. In the analysis of scRNA-Seq data, both gene level filtering and cell level filtering are critical for clustering regardless of which clustering method to use. We recommend to rank genes by their variations among all cells and choose top 500-1,000 highly variable genes. In addition, we also recommend to run DIMM-SC 5~10 times, each with different random seeds, and choose the one with the largest likelihood as the final results. For the number of clusters, we can pre-define it based on prior knowledge on the tissue or determine it using some model checking criteria such as AIC or BIC (Akaike, 1974; Schwarz, 1978). As shown in **Figure S5**, AIC and BIC work well in the analysis of simulated datasets, the performance in real data needs further exploration. Alternatively, it can be determined using the procedure described in ADPcluster (Wang and Xu, 2015) or the Dirichlet process (Teh, 2011). DIMM-SC is currently implemented in R with satisfactory computing efficiency for most needs so far. We will further speed it up using a C++/Rcpp implementation to accommodate larger-scaled data.

There are several noticeable limitations of our method. First, DIMM-SC only models variations among different cells from one single individual. To jointly model scRNA-Seq data from multiple individuals, a hierarchical structure can be posed in the current method to account for the individual level heterogeneity, but a more sophisticated numerical algorithm will be needed to reduce the computational cost. Second, DIMM-SC is an unsupervised method that infers structures from all data. Prior knowledge on cell-type-specific biomarkers may further improve the clustering accuracy. To use such prior information, a semi-supervised approach is needed to guide cluster inference. Furthermore, existing scRNA-Seq data from purified cells (e.g. via flow cytometry) can serve as external reference panels or training datasets to reduce experimental biases, remove outliers, and improve clustering reliability. Last but not least, our DIMM-SC model ignores the measurement errors and uncertainties buried in the UMI count matrix. Multiple factors such as drop-out event, mapping percentage, sequencing depth, and PCR efficiency are not considered in the current model. These limitations can be largely overcome by extending our method. We will explore these directions in the near future.

We noticed that similar models have been proposed in the field of text-mining (Yamamoto and Sadamitsu, 2005) and microbiome (Holmes, et al., 2012), where word, article, and topic or taxa, individual, and meta-community are studied. However, in those applications, the clusters are not well defined and require a careful interpretation. On the contrary, scRNA-Seq data usually consist of a set of known cell types from prior knowledge and have a much larger signal-to-noise ratio for the clustering analysis. Although sharing the common types of data structure, these fields have different fundamental questions, so existing methods proposed from other fields need to be tailored or extended to incorporate intrinsic characteristics of scRNA-Seq data. For example, CellTree adapts the LDA approach from the text-mining field. Although LDA is more flexible and more widely used in text-mining field than the Dirichlet mixture model based methods, we have showed that DIMM-SC is more accurate, stable and efficient than CellTree in both simulation studies and real data applications in the context of scRNA-Seq clustering analysis.

In summary, we provide a novel statistical method and an efficient computational tool DIMM-SC for clustering droplet-based single cell transcriptomic data, which facilitates rigorous statistical inference of cell population heterogeneity. We are confident that DIMM-SC will be highly useful for the fast-growing community of large-scale single cell transcriptome analysis.

## Acknowledgements

The authors would like to thank Dr. Zhiguang Huo and Dr. Kong Chen for providing valuable insights during the initial stage of this project.

## Funding

This work is supported by National Institute of Health grants R01HG007358 (W.C.), Children's Hospital of Pittsburgh (W.C. and Z.S.), U54DK107977 (M.H.), and the National Science Foundation of China Grant 11401338 (K.D.).

*Conflict of Interest:* none declared.

## Supplemental materials

### 1. The E-M algorithm

We used the E-M algorithm to maximize the log posterior distribution. Specifically, we first denoted $P\left(z_j = k\right) = \pi_k$, where $\pi_k$ is the proportion of the $k$ th cell type among all cells. We then treated $z_j$ as missing data and used the E-M algorithm to estimate $\alpha_{1k}, \alpha_{2k}, \dots, \alpha_{Gk}$ and $\pi_k$. The complete data likelihood is:

$$\prod_{j=1}^{C} P(\boldsymbol{x_j}, z_j) = \prod_{j=1}^{C} \left\{ \left( \prod_{i=1}^{G} \frac{\Gamma(x_i + \alpha_{ik})}{\Gamma(\alpha_{ik})} \right) \frac{\Gamma(|\boldsymbol{\alpha}_{(k)}|)}{\Gamma(T_j + |\boldsymbol{\alpha}_{(k)}|)} \right\}^{I(z_j=k)},$$

where $\left|\boldsymbol{\alpha}_{(k)}\right| = \alpha_{1k} + \alpha_{2k} + \cdots + \alpha_{Gk}$ and the log likelihood is:

$$\log \prod_{j=1}^{C} P(\boldsymbol{x_j}, z_j) = \sum_{j=1}^{C} I(z_j = k) \log \left\{ \left( \prod_{i=1}^{G} \frac{\Gamma\left(x_{ij} + \alpha_{ik}\right)}{\Gamma(\alpha_{ik})} \right) \frac{\Gamma(|\boldsymbol{\alpha}_{(k)}|)}{\Gamma(T_j + |\boldsymbol{\alpha}_{(k)}|)} \right\}.$$

E-step:

At the $t$ th iteration, with the current realization of parameters $\Theta^{(t)} = (\alpha_{1k}^{(t)}, \alpha_{2k}^{(t)}, \dots, \alpha_{Gk}^{(t)},\ \pi_k^{(t)})$, the conditional expectation is:

$$E_{z_j|\boldsymbol{x_j},\Theta^{(t)}} \log P\left(\boldsymbol{x_j}, z_j\right) = \log \left\{ \left( \prod_{i=1}^{G} \frac{\Gamma\left(x_{ij} + \alpha_{ik}\right)}{\Gamma(\alpha_{ik})} \right) \frac{\Gamma\left(\left|\boldsymbol{\alpha}_{(k)}\right|\right)}{\Gamma\left(T_j + \left|\boldsymbol{\alpha}_{(k)}\right|\right)} \right\} * P\left(z_j = k \middle| \boldsymbol{x_j}, \Theta^{(t)}\right),$$

where

$$P\left(z_j = k \middle| \boldsymbol{x_j}, \Theta^{(t)}\right) = \frac{\left( \prod_{i=1}^{G} \frac{\Gamma\left(x_{ij} + \alpha_{ik}^{(t)}\right)}{\Gamma\left(\alpha_{ik}^{(T)}\right)} \right) \frac{\Gamma\left(\left|\boldsymbol{\alpha}_{(k)}^{(t)}\right|\right)}{\Gamma\left(T_j + \left|\boldsymbol{\alpha}_{(k)}^{(t)}\right|\right)} \pi_k^{(t)}}{\sum_{k=1}^{K} \left( \prod_{i=1}^{G} \frac{\Gamma\left(x_{ij} + \alpha_{ik}^{(t)}\right)}{\Gamma\left(\alpha_{ik}^{(T)}\right)} \right) \frac{\Gamma\left(\left|\boldsymbol{\alpha}_{(k)}^{(t)}\right|\right)}{\Gamma\left(T_j + \left|\boldsymbol{\alpha}_{(k)}^{(t)}\right|\right)} \pi_k^{(t)}} = \delta_{jk}.$$

Here $\delta_{jk}$ represents the probability that the $j$ th cell belongs to the $k$ th cluster. We calculated $\delta_{jk}$ in the E-step at each iteration.

M-step:

At the $t$ th iteration, the estimation of the proportion of the $k$ th cell type is $\hat{\pi}_k^{(t+1)} = \sum_{j=1}^{C} \delta_{jk}{}^{(t)}/C$. The update formula for $\alpha_{1k}, \alpha_{2k}, \dots, \alpha_{Gk}$ is derived from the Minka's fixed-point iteration for the leaving-one-out (LOO) likelihood (Minka , 2000):

$$\hat{\alpha}_{ik}^{(t+1)} = \alpha_{ik}^{(t)} \frac{\sum_{j=1}^{C} \delta_{jk}\left\{x_{ij}/(x_{ij} - 1 + \alpha_{ik}^{(t)})\right\}}{\sum_{j=1}^{C} \delta_{jk}\left\{T_j/(T_j - 1 + |\boldsymbol{\alpha}_{(k)}^{(t)}|)\right\}}.$$

After the M-step, we calculated $\sum_{k=1}^{K}(\hat{\pi}_k^{(t+1)} - \hat{\pi}_k^{(t)})^2$ and the relative difference of log likelihood between two consecutive iterations. Convergence tolerances for difference between iterations are pre-defined. We repeated the above steps until the convergence of log likelihood and $\hat{\pi}_k^{(t)}$, or a maximum number of iterations was reached. The default maximum number of iterations is 200.

**Figure S1. The t-SNE projection of the simple case.**

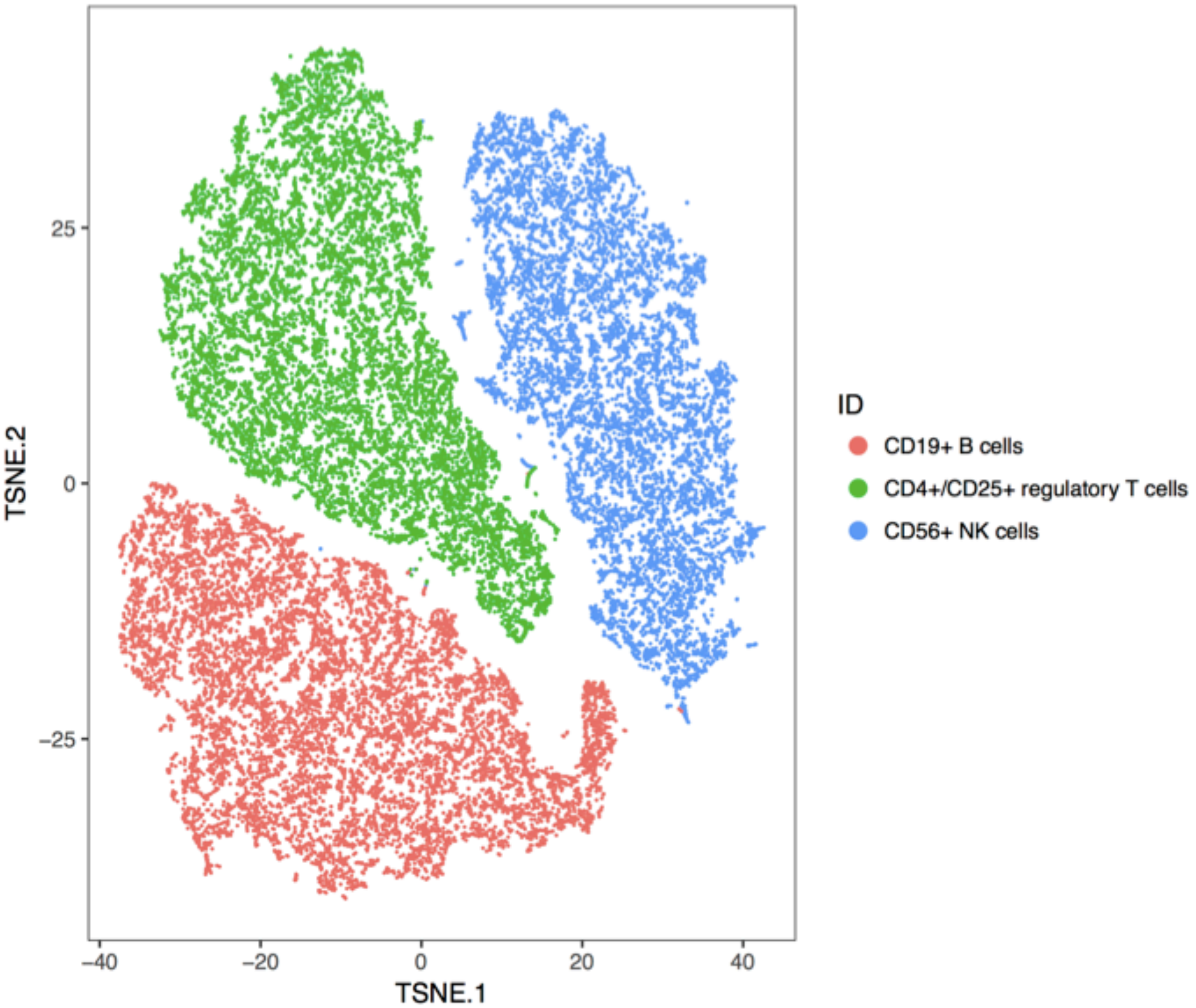

**Figure S2. The t-SNE projection of the challenging case.**

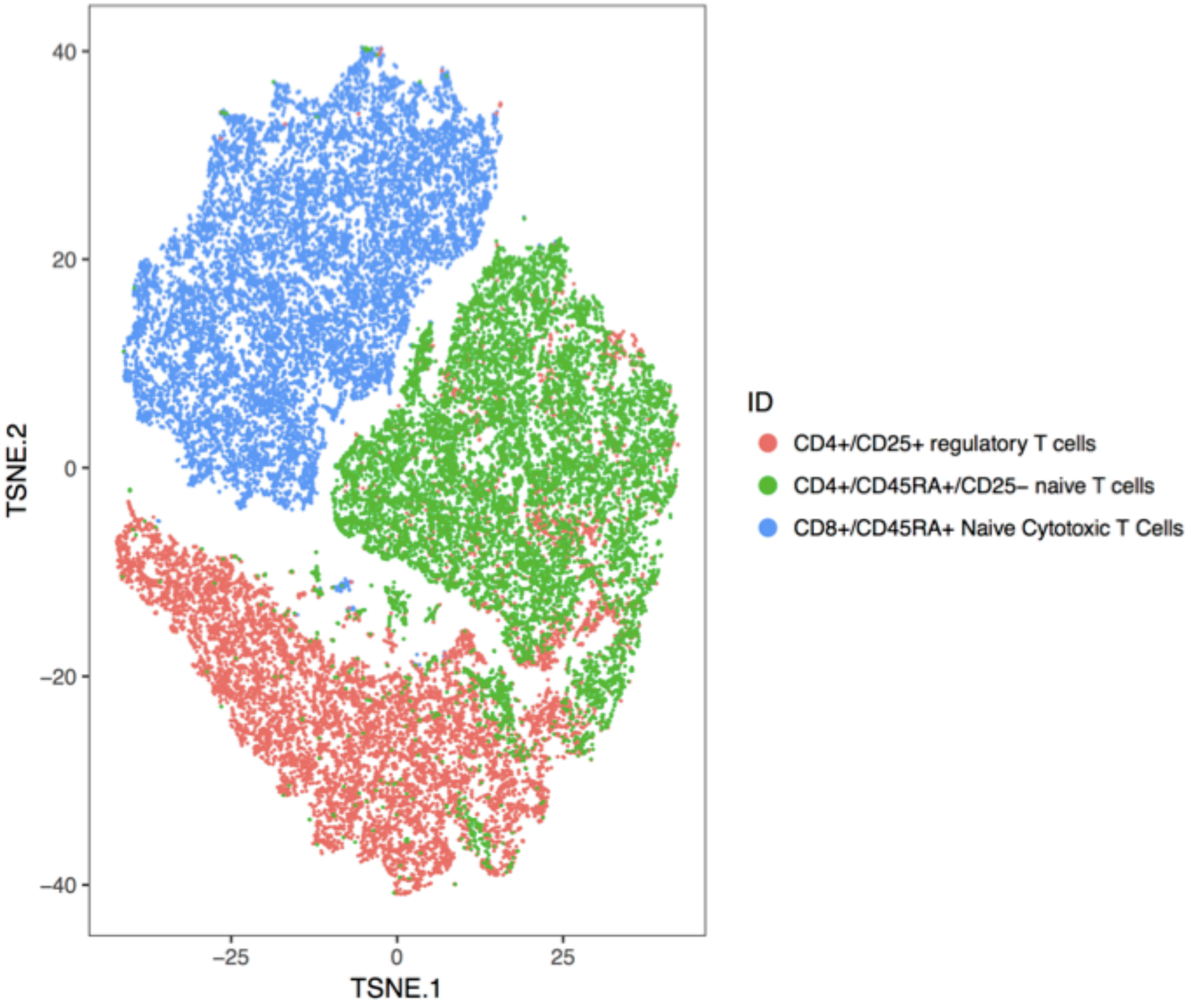

**Figure S3. The boxplots of ARI for seven clustering methods across 50 simulations in the challenging case.**

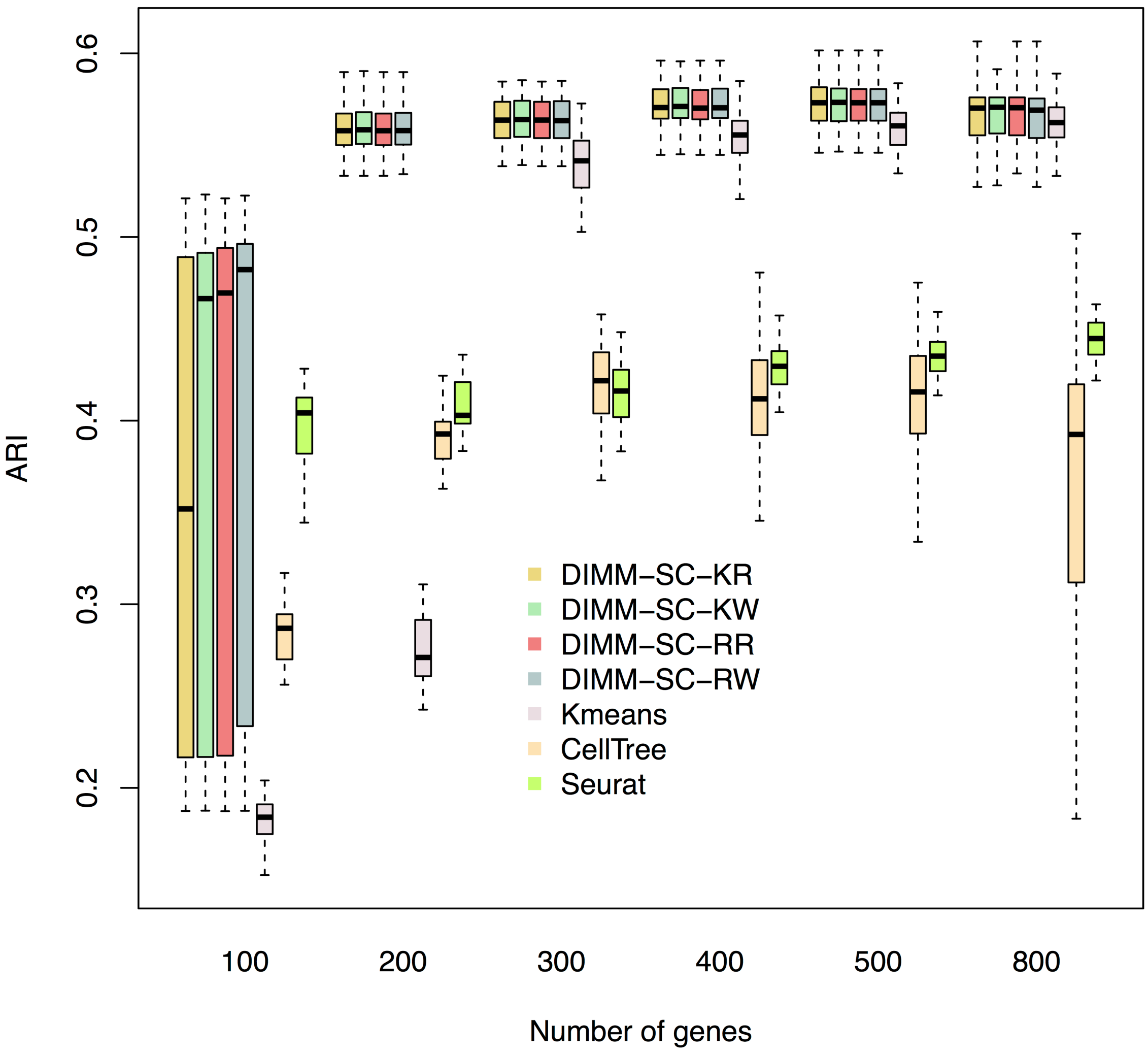

**Figure S4A. The histogram of proportion** $p_i$ **for gene *RPS27* and gene *RPL18A* and the theoretical marginal beta distribution (solid blue line) in CD56+ NK cells.** To obtain the theoretical marginal beta distribution, we used the top 1% (327) genes and calculate $\alpha_i$ of the Dirichlet distribution by the Ronning's method (Ronning, 1989).

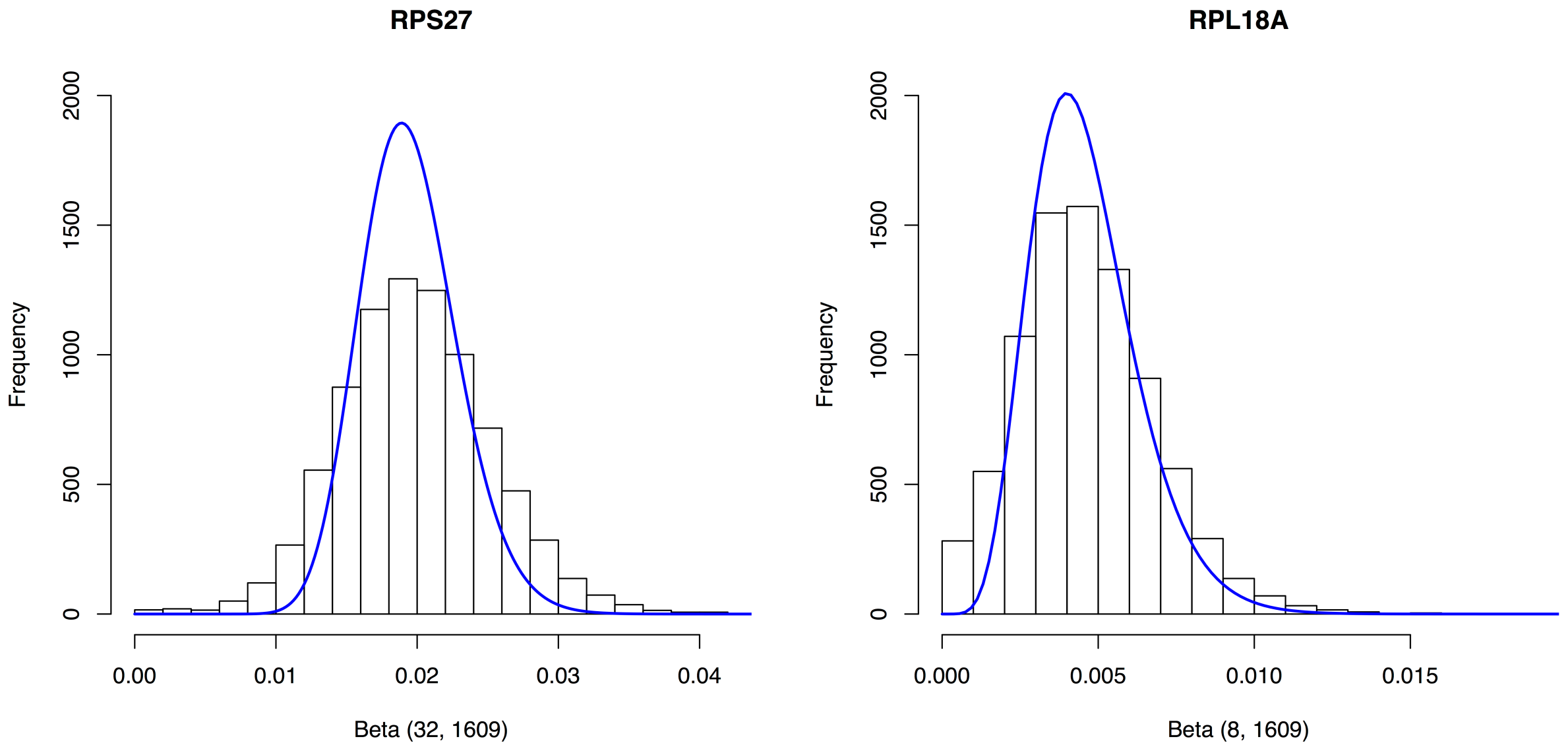

**Figure S4B. The scatter plot of the log mean of $p_i$ versus the log variance of $p_i$ in the CD56+ NK cells.** α and β are linear regression intercept and slope, respectively. Each dot represents one gene. This figure includes the top 1% (327) highly variable genes.

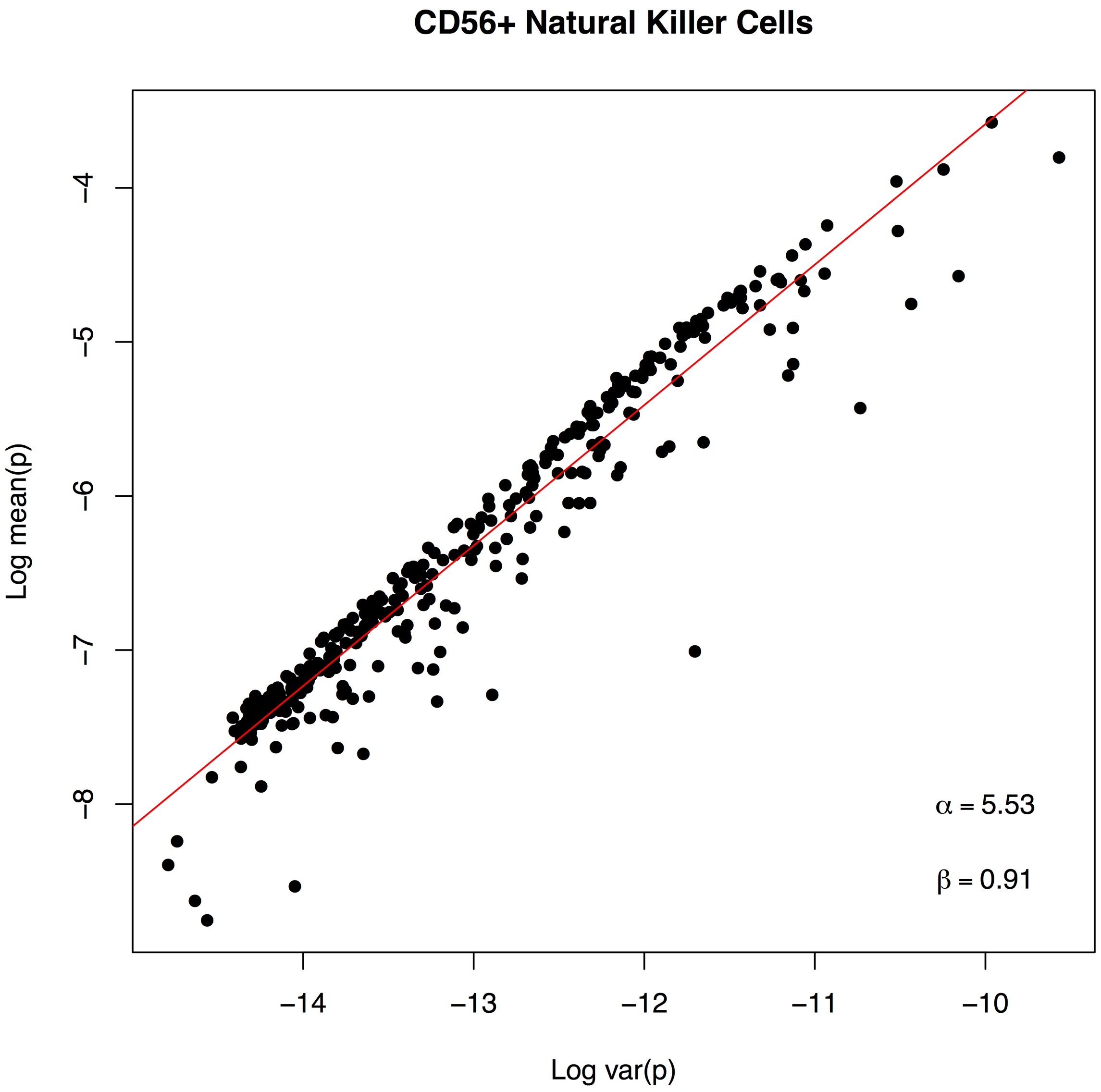

**Figure S5. The dot plots of AIC and BIC for the final clustering results in the simulated dataset, where the true number of clusters is 3.** Blue dots and red dots denote values of BIC and AIC, respectively. Black dots denote ARIs.

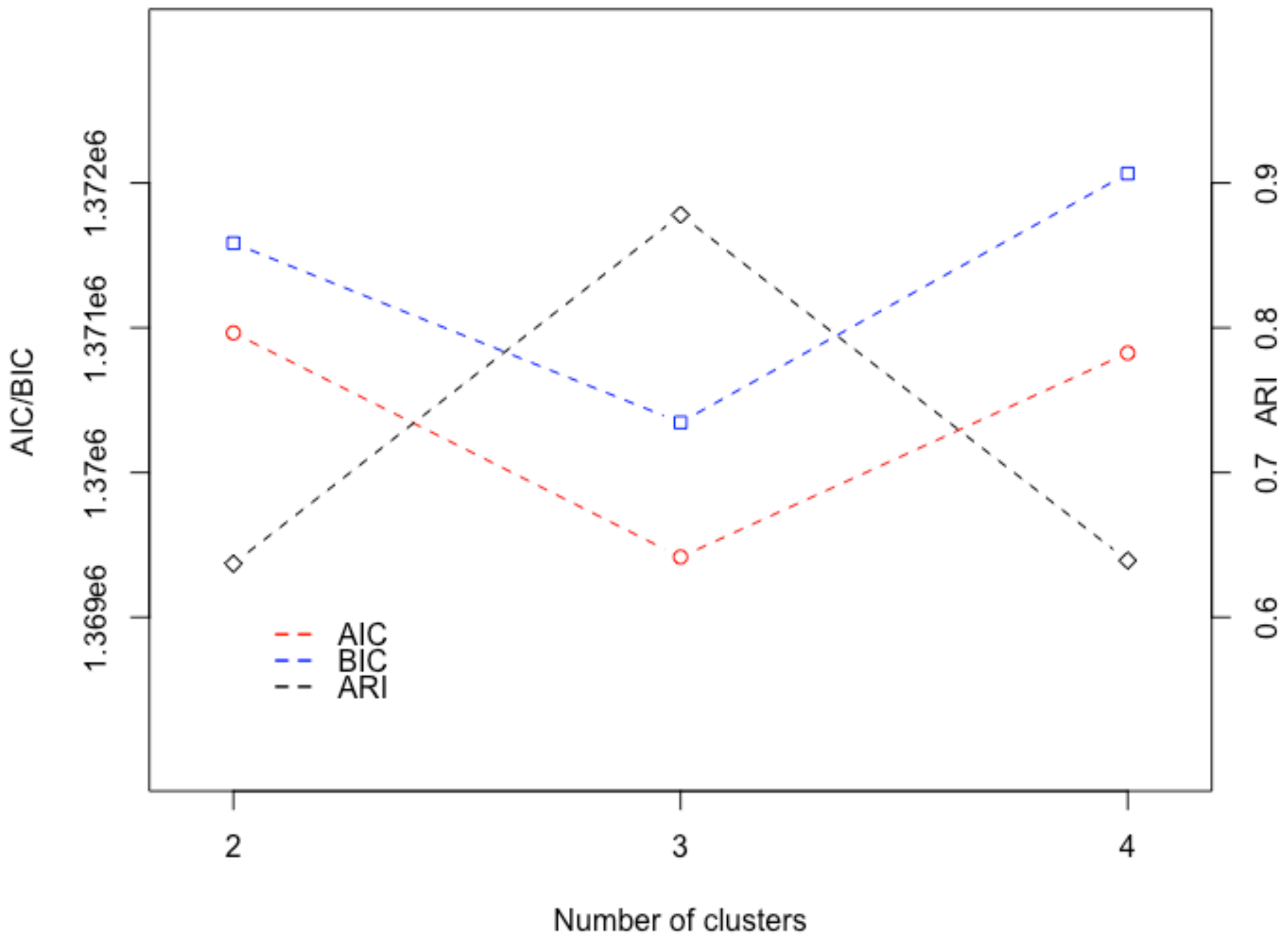